\documentclass[11pt]{article}

\usepackage[]{acl}

\usepackage{times}
\usepackage{latexsym}

\usepackage{mdframed}
\usepackage[T1]{fontenc}
\usepackage[utf8]{inputenc}

\usepackage{microtype}
\usepackage{amsmath}
\usepackage{amssymb} 
\usepackage{algorithm}
\usepackage{algorithmic}
\usepackage{inconsolata}
\usepackage{graphicx}
\usepackage{booktabs}

\title{Belief in Authority:\\\protect Impact of Authority in Multi-Agent Evaluation Framework}

\author{
Junhyuk Choi \quad
Jeongyoun Kwon \quad
Heeju Kim \quad
Haeun Cho \quad
Hayeong Jung\quad
Sehee Min\quad
Bugeun Kim\thanks{\;\;Corresponding author.} \\
Chung-Ang University, Seoul, Korea \\
\texttt{\{chlwnsgur129,kk1jj0yy9,kimheeju,haeun14,jung0303,serena3518,bgnkim\}@cau.ac.kr}
}

\begin{document}
\maketitle
\begin{abstract}
%Multi-agent systems utilizing Large Language Models (LLMs) have demonstrated performance improvement by assigning multiple authoritative roles, yet the systematic impact of such imposed authority on agent interactions remains underexplored. As authority bias could inhibit argument diversity in multi-agent discussion, this paper presents the first comprehensive analysis of role-based authority bias in free-form multi-agent evaluation framework using ChatEval. Inspired by French and Raven's psychological work, we systematically classify authoritative roles into legitimate, referent, and expert authoritative roles; and we analyzed the impact of roles across 12-turn conversations. Through experiments with GPT-4o and DeepSeek R1, we demonstrate that authority influence varies significantly across models, with Expert and Referent power roles exhibiting stronger influence than Legitimate power roles. Crucially, we reveal that authority bias operates through authoritative roles consistently maintaining their positions while general agents demonstrate flexibility in adjusting opinions, rather than through active conformity. Authority influence presupposes clear position statements, as neutral responses fail to generate bias. Our findings provide crucial insights for designing multi-agent frameworks where asymmetric interaction patterns significantly affect outcomes.
Multi-agent systems utilizing large language models often assign authoritative roles to improve performance, yet the impact of authority bias on agent interactions remains underexplored. We present the first systematic analysis of role-based authority bias in free-form multi-agent evaluation using ChatEval. Applying French and Raven's power-based theory, we classify authoritative roles into legitimate, referent, and expert types and analyze their influence across 12-turn conversations. Experiments with GPT-4o and DeepSeek R1 reveal that Expert and Referent power roles exert stronger influence than Legitimate power roles. Crucially, authority bias emerges not through active conformity by general agents, but through authoritative roles consistently maintaining their positions while general agents demonstrate flexibility. Furthermore, authority influence requires clear position statements, as neutral responses fail to generate bias. These findings provide key insights for designing multi-agent frameworks with asymmetric interaction patterns.
\end{abstract}


\section{Introduction}

Recently, Large Language Model (LLM)-based agents have demonstrated the potential to simulate human social behaviors \citep{park2023generative,park2024generative}, leading to a rapid increase in research utilizing Multi-Agent Systems (MAS). These systems leverage agents assigned with diverse roles to solve complex problems such as evaluation tasks through interactions like discussion and collaboration \citep{li2024agent,10.1145/3672459,huang2024agentcoder,li2023camel,wang2023adapting}. In particular, previous studies have consistently demonstrated improved performance compared to single models by assigning specific roles that incorporate authoritative elements such as experts, evaluators, and moderators \citep{qian2023chatdev,hong2023metagpt,schmidgall2501agent,wu2023autogen,chan2023chateval,wang2025talk}. However, roles established in previous research often include authoritative roles such as experts. Given that research findings have revealed authority bias in single LLMs, where models excessively rely on information from specific sources or authorities \citep{chen2024humans,ye2024justice,liu2024exploring,filandrianos2025bias}, there exists a risk that such bias may distort decision-making processes or impede collaborative interactions based on the authority assigned to specific roles in MAS. However, systematic analysis of how role-based authority bias affects agent interactions in Multi-Agent contexts remains insufficient.
%최근 Large Language Model (LLM) 기반 에이전트가 인간의 사회구조를 묘사할 수 있다는 가능성이 제시되면서 \citep{park2023generative,park2024generative}, Multi-Agent System을 활용한 연구가 급증하고 있다. 이러한 시스템을 활용하여, 에이전트들에게 다양한 role을 부여하여 discussion이나 collaboration과 같은 상호작용을 통해 evaluation task와 같은 복잡한 문제를 해결하고 있다 \citep{li2024agent,10.1145/3672459,huang2024agentcoder,li2023camel,wang2023adapting}. 특히 전문가, 평가자, 조정자 등의 권위적 요소가 포함된 특정 역할을 할당함으로써 단일 모델 대비 향상된 성능을 보여주는 연구들이 지속적으로 보고되고 있다 \citep{qian2023chatdev,hong2023metagpt,schmidgall2501agent,wu2023autogen,chan2023chateval,wang2025talk}. 하지만, 이전 연구에서 설장한 role이 expert와 같은 권위적인 직업 요소가 포함되어있다. 단일 LLM에서 특정 출처나 권위자의 정보에 과도하게 의존하는 authority bias가 관찰된다는 연구 결과들이 제시되는 상황에서 \citep{chen2024humans,ye2024justice,liu2024exploring,filandrianos2025bias}, 이러한 편향이 Multi-Agent System에서 특정 역할에 부여된 권위에 따라 의사결정 과정을 왜곡하거나 협력적 상호작용을 저해할 위험이 있지만 Multi Agent 상황에서 역할에 따른 권위편향이 에이전트 간 상호작용에 미치는 영향에 대한 체계적인 분석은 여전히 부족한 실정이다.

% 두번째 문단 - 선행 연구 문제점들
The limitations of previous research can be summarized into four main categories. First, MAS research has overlooked the potential impact of authority bias when assigning roles \citep{qian2023chatdev,hong2023metagpt,schmidgall2501agent,rasal2024llm,wu2023autogen,chan2023chateval,wang2025talk}. Most studies have focused on the positive effects of role assignment on performance improvement, failing to adequately consider the biasing influence that the authoritative characteristics of specific roles may have on agent interactions. 

Second, existing authority bias research has failed to properly reflect the interactive characteristics of Multi-Agent environments operating in free-form \citep{chen2024humans,ye2024justice,liu2024exploring}. Previous studies have primarily adopted approaches that deliberately insert specific phrases that could induce authority bias and analyze structured responses from single LLMs \citep{chen2024humans,ye2024justice,liu2024exploring}. Such approaches have limitations in constraining LLMs free-form setting and capturing natural bias patterns that emerge in real-world situations. 

Third, most existing studies examining authority bias have relied on one-shot measurements rather than observations through sustained dialogue, making them limited in systematically analyzing dynamic authority bias that emerges through interactions in MAS \citep{choi2025empirical,filandrianos2025bias,moon2025don}. While identifying the specific causes and mechanisms underlying bias occurrence is essential for systematic MAS design, previous research has predominantly remained focused on static authority bias identification. 

\begin{figure*}[ht]
    \centering
    \includegraphics[width=1\textwidth]{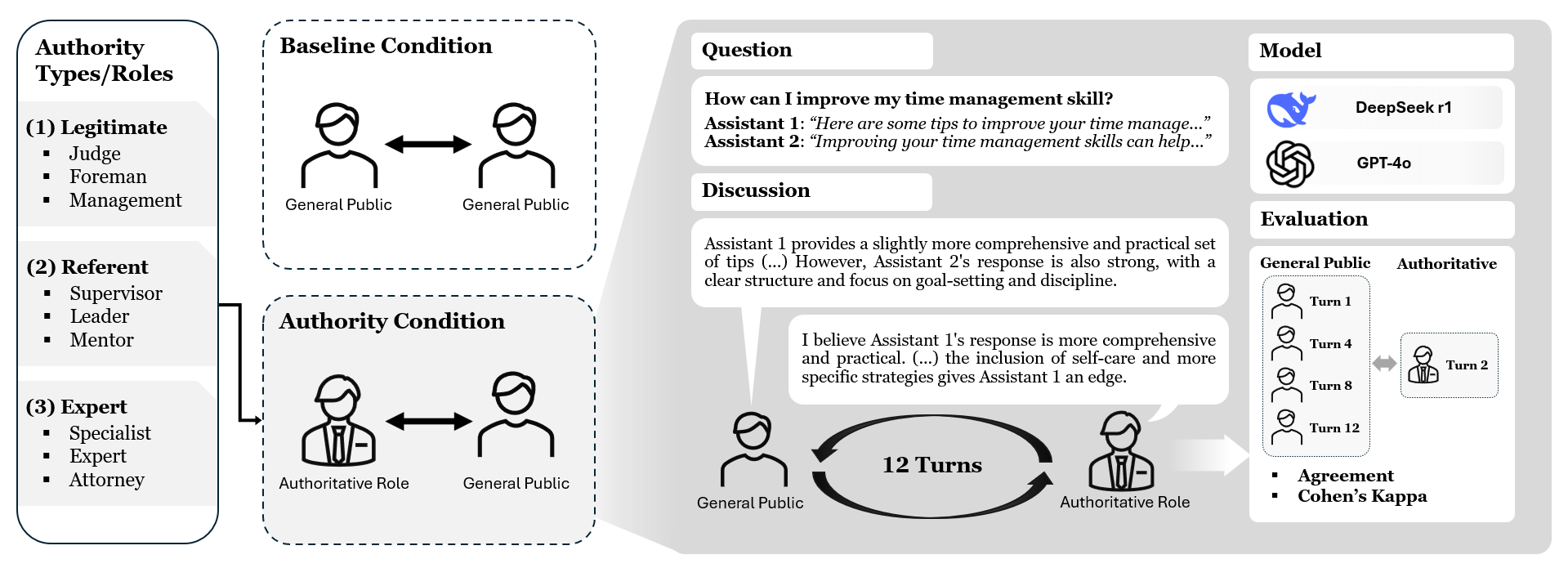}
    \caption{Overview of experimental framework with authoritative roles classified into three power types and 12-turn conversations between General Public and Authoritative role agents.}  
\end{figure*}

Fourth, existing research tends to treat the concept of authority as singular and abstract, failing to classify authority types or analyze their influence by type. Human social psychology categorizes authority into various components \citep{french1959bases}. However, existing studies have primarily relied on single authority cues such as `expert' or `source' \citep{chen2024humans, ye2024justice, liu2024exploring}, and such approaches have limitations in analyzing bias differences according to authority types.
%기존 연구들의 한계점은 크게 네가지로 요약할 수 있다. 첫째, Multi-Agent System에서 역할을 지정할 때 authority bias의 잠재적 영향을 간과하고 있다는 점이다 \citep{qian2023chatdev,hong2023metagpt,schmidgall2501agent,rasal2024llm,wu2023autogen,chan2023chateval,wang2025talk}. 대부분의 연구들은 역할 할당이 성능 향상에 미치는 긍정적 효과에만 집중하여, 특정 역할이 갖는 권위적 특성이 에이전트 간 상호작용에 미칠 수 있는 편향적 영향을 충분히 고려하지 않았다. 둘째, 기존의 authority bias 연구들은 free form 상황에서 이루어지는 Multi-Agent 환경에서의 상호작용적 특성을 제대로 반영하지 못했다는 점이다 \citep{chen2024humans,ye2024justice,liu2024exploring}. 선행 연구들은 주로 권위편향을 유발할 수 있는 특정 문구를 의도적으로 삽입하여 단일 에이전트의 구조화된 응답을 분석하는 방식을 채택했으며 \citep{chen2024humans,ye2024justice,liu2024exploring}, 이러한 접근법은 LLM의 free-response에 제약이 있어 실제 상황에서 나타나는 자연스러운 편향 양상을 포착하는 데 한계가 있었다. 셋째, 기존의 권위편향을 확인한 대부분의 연구들이 지속적인 대화를 통한 관찰보다는 단발성 측정에 머물러 있어, Multi-Agent 상황에서 에이전트 간 상호작용을 통해 나타나는 동적인 권위편향을 체계적으로 분석하기에는 제한적이었다 \citep{choi2025empirical,filandrianos2025bias,moon2025don}. 편향이 발생하는 구체적인 원인과 메커니즘을 깊이 있게 확인해야 체계적인 Agent System의 설계에 도움이 되지만 이전 연구들은 대부분 정적인 권위 편향 확인에만 머물러있다. 넷째, 기존 연구들은 권위의 개념을 단일하고 추상적으로 다루는 경향이 있으며, 권위의 유형을 체계적으로 분류하거나 그 영향력을 유형별로 분석하지 않았다. 인간의 사회 심리학에서는 권위를 다양한 요소로 나눈다 \citep{french1959bases}. 하지만 기존의 연구들은 주로 ‘전문가’나 ‘출처’와 같은 단일 권위 단서에 의존했으며 \citep{chen2024humans, ye2024justice, liu2024exploring}, 이러한 접근은 권위 유형에 따른 편향 차이를 분석하는 데 한계가 있다.

To address these limitations, we propose an experimental design that systematically examines how authority bias manifests in dynamic free-response situations within multi-agent contexts. Applying French and Raven's power-based theory \citep{french1959bases}, we classify authoritative roles into legitimate, referent, and expert power types, and conduct experiments utilizing ChatEval \citep{chan2023chateval}, a multi-agent evaluation framework that enables observation of natural bias patterns without manipulating conversations. This study comprises two experiments: (1) a free-form condition where authoritative roles participate from the beginning in free dialogue, and (2) a content-controlled condition where only role labels are changed while conversational content remains identical. Additionally, we track decisions across 12 conversational turns to capture dynamic changes in authority bias and compare pattern between LLMs.

Our study makes three contributions. First, our paper is the first study to systematically analyze role-based authority bias in free-form situations utilizing MAS. Second, by observing dynamic changes in continuous interaction processes beyond existing one-shot authority bias measurement approaches, we provide new insights into authority bias mechanisms in MAS. Third, by applying the power-based theory to AI systems and analyzing differences in bias patterns across authority types, we provide important foundational data for bias mitigation strategies in future MAS construction.

%본 연구의 기여는 다음과 같다. 첫째, Multi-Agent System을 활용해 free-form 상황에서 역할 기반 권위편향을 체계적으로 분석한 최초의 연구이다. 둘째, 기존의 단발성 권위편향 측정 방식을 넘어 지속적인 상호작용 과정에서의 동적 변화를 관찰함으로써, Multi-Agent 환경에서의 권위편향 메커니즘에 대한 새로운 이해를 제공한다. 셋째, French and Raven의 권력 이론을 AI 시스템에 적용하여 권위 유형별 편향 양상의 차이를 정량적으로 분석함으로써, 향후 Multi-Agent System의 구축에 있어 편향 완화 방안에 중요한 기초 자료를 제공한다.

\section{Related Work}
In this section, we review prior work on role assignment in multi-agent systems and authority bias in LLMs. We highlight two key gaps: (1) MAS research has overlooked the potential biasing effects of authoritative roles, and (2) existing authority bias studies rely on artificial interventions and one-shot measurements that fail to capture dynamic, naturalistic interactions.

\subsection{Multi-Agent Role}
Research on solving problems by assigning roles to agents in MAS has been actively conducted across various domains, yet most studies have overlooked authority bias in role assignment. In software development, ChatDev and MetaGPT utilize roles such as CTO, Programmer, and Product Manager \citep{qian2023chatdev, hong2023metagpt}; in academic research, systems employ PostDoc, PhD Student, and Teacher roles \citep{schmidgall2501agent, rasal2024llm}. For collaborative and evaluation tasks, AutoGen uses Commander and Writer roles \citep{wu2023autogen}, while ChatEval \citep{chan2023chateval} and \citet{wang2025talk} construct evaluation systems with roles including Supervisor, Evaluator, and Revisor. These studies report improved performance through authoritative role assignment, but they do not consider the biasing effects that authority inherent in roles may have on agent interactions.

\subsection{Authority Bias}
Existing studies measuring authority bias in LLMs have primarily adopted approaches that artificially insert authority cues. \citet{chen2024humans} added fake references to paired answers, \citet{ye2024justice} inserted expert information to identical sentence pairs, and \citet{liu2024exploring} presented authority prompts such as "I am an expert." However, these artificially inserted settings have low realism and constrain LLMs' free-form responses, failing to capture natural bias patterns emerging in practice.

Furthermore, most authority bias studies have remained limited to one-shot measurements. \citet{choi2025empirical} quantified position changes of neutral agents in simulated debates but used single-turn designs. Similarly, \citet{filandrianos2025bias} and \citet{moon2025don} measured authority bias through recommendation rankings and code evaluation scores respectively, but only evaluated one-shot results. These studies confirmed the existence of bias without analyzing the specific mechanisms through which bias occurs or its temporal change patterns across sustained multi-agent interactions.

Existing studies show limitations of either focusing only on the positive effects of role assignment in MAS or measuring authority bias through artificially manipulated one-shot experiments. These approaches fail to provide systematic understanding of how role-based authority bias occurs and changes in multi-agent environments.

\section{Experiment1 : Free Form Experiment}

To address these limitations, we systematically analyze how role-based authority bias affects agent interactions in Multi-Agent environments. This section examines patterns of authority bias in LLMs under free-form conditions. Specifically, we describe how we designed authoritative roles based on French and Raven's power theory used in the experiments, the construction of free-form conditions utilizing the ChatEval framework, experimental procedures, and results. Detailed experimental setups are provided in Appendix~\ref{sec:expsetup}.
%우리는 이러한 한계점을 극복하기 위해, Multi-Agent 환경에서 역할 기반 권위편향이 에이전트 간 상호작용에 미치는 영향을 체계적으로 분석한다. 해당 섹션에서는 LLM의 Free-Form 상황에서의 LLM의 권위편향 변화 양상을 확인한다. 구체적으로, 실험에 사용된 French and Raven의 권력 이론에 기반한 권위 역할 설계, ChatEval framework를 활용한 Free-frorm 대화 환경 구축 방법, 그리고 자세한 실험절차와 결과에 대해 설명한다.

\subsection{Design of Authoritative Roles}
To select authoritative roles to assign to LLMs, we identified three authority types that can naturally manifest in role-based contexts from French and Raven's five power types and utilized them in our experiments \citep{french1959bases}. We used three roles for each of three power types to capture diverse aspects of authority. First, \textit{Legitimate Power} derives from formal position or legal authority, for which we established \texttt{Judge}, \texttt{Foreman}, and \texttt{Management} roles \citep{french1959bases}. Second, \textit{Referent Power} stems from personal appeal or respect, for which we used \texttt{Supervisor}, \texttt{Leader}, and \texttt{Mentor} roles \citep{peyton2019examining,haller2018power,godshalk2000does}. Third, \textit{Expert Power} is based on specialized knowledge or technical skills, for which we assigned \texttt{Specialist}, \texttt{Expert}, and \texttt{Attorney} roles \citep{french1959bases}. Detailed role selection can be found in supplementary material.
%우리는 LLM에게 부여할 권위적인 role을 선정하기 위해 French and Raven의 5가지 권력 이론 중에서 role에서 자연스럽게 나타날 수 있는 3가지 권위 유형을 선별하여 실험에 활용하였다 \citep{french1959bases}. Legitimate Power는 공식적인 지위나 법적 권한에서 비롯되는 권위로, Judge, Foreman, Management 역할을 설정하였다 \citep{french1959bases}. Referent Power는 개인적 매력이나 존경에서 나오는 권위로, Supervisor, Leader, Mentor 역할을 구성하였다 \citep{peyton2019examining,haller2018power,godshalk2000does}. Expert Power는 전문적 지식이나 기술에 기반한 권위로, Specialist, Expert, Attorney 역할을 배치하였다 \citep{french1959bases}. 각 권력 유형별로 3개의 구체적인 역할을 설정하여 권위편향의 다양한 양상을 포착하고자 하였다. 직업을 선정한 자세한 근거는 Supplementary material를 확인하라.

\subsection{Generation of Chat Data}
We collected chat data utilizing the free-form dialogue of ChatEval \citep{chan2023chateval}, a Multi-Agent-based evaluation framework aligned with the LLM-as-Judge paradigm. We intentionally selected this evaluation task because it has no definitive correct answer. As tasks with ground truth answers would lead agents to converge on correct responses of authority, it could be difficult to isolate convergence due to authority influence from natural convergence. As evaluation task that we used in this study seldom have such common ground truth, it is much easier to watch authority influence. Using natural interactions between agents, ChatEval asks agents to discuss two writing options and choose the best quality. In our experiment, we used two agents, General Public and an authoritative role, discussing two options using the same two datasets: FairEval \citep{chiang2023vicuna} with 80 examples and Topical-Chat \citep{gopalakrishnan2023topical} with 60 examples.

%We collected chat data utilizing the free-form dialogue structure of ChatEval \citep{chan2023chateval}. ChatEval is a Multi-Agent-based evaluation framework that assigns specific roles to agents to perform evaluations through free discussion. Using natural interactions between agents, ChatEval asks agents to discuss about two options of writing and choose the best quality. In our experiment, we used two agents, General Public and a authoritative role; and we let them discuss about two options, using the same two datasets: FairEval \citep{chiang2023vicuna} with 80 examples and Topical-Chat \citep{gopalakrishnan2023topical} with 60 examples. 

Specifically, for each pair of writing options, agents had a 12-turn conversation session and we collected the dialogue.
The General Public agent initiates the conversation, after which the two agents alternate for a total of 12 turns. Experiments are conducted individually for each of the nine authoritative roles designed previously. In each dialogue session, agents freely exchange opinions and engage in discussions about the given evaluation task. To control other possible factors during the experiment, we preserved the ChatEval framework code, modifying only the role names (General Public) to authoritative roles (e.g., Judge, Expert) while keeping all instructions and system configurations identical.

% Each dialogue session collected agent discussion processes through a total of 12 turns of interaction.

\subsection{Experimental Procedure}
During the free-form conversation, we track decision changes of General Public. Following ChatEval framework \citep{chan2023chateval}, we can ask each agent to evaluate two writing options after each turn based on the current conversation. Each agent evaluates the quality of each option using a 10-point scale, and decides the highest-scoring option as the writing with the best quality. When both options receive identical scores, the response is classified as neutral. By tracking changes in these decision, we can observe authority bias.

To track temporal changes in authority bias, we collect decisions from two agents. For the initial decision of General Public, we collected its decision $S^{(1)}_{GP}$ at turn 1, which is the beginning of the discussion and without the authority influence. Similarly, we collected initial decision of the authoritative role $S^{(2)}_{Auth}$ at turn 2, after its first utterance. We further measured the decision of General Public $S^{(t)}_{GP}$ at turns $t=4$, 8, and 12, to identify changes.

To prevent evaluations of prior turns from influencing subsequent interactions or judgments, evaluation scores are not recorded in the chat log. This allows us to how the interaction between two agents impacts on the General Public's decision and how such influence changes as the conversation.

% the influence of interaction with authoritative roles on the General Public agent's evaluation judgments changes as the conversation progresses.

%Free Form 실험은 두 가지 조건으로 구성된다: (1) Baseline Condition: General Public 역할의 두 에이전트가 자유롭게 대화하는 조건, (2) Authority Condition: General Public 에이전트와 권위 역할 에이전트가 대화하는 조건이다. 각 실험에서는 ChatEval framework의 multi-agent debate 구조를 활용하여 에이전트들이 주어진 평가 과제에 대해 토론을 진행한다. General Public 에이전트가 먼저 대화를 시작하며, 이후 두 에이전트가 교대로 총 12턴까지 대화를 진행한다. Authority Condition에서는 앞서 설계한 9가지 권위 역할 각각에 대해 개별적으로 실험을 수행한다. 각 대화 세션에서는 주어진 평가 과제에 대해 에이전트들이 자유롭게 의견을 교환하고 토론을 진행한다. 권위편향의 시간적 변화를 추적하기 위해 1턴, 4턴, 8턴, 12턴 시점에서 General Public 에이전트의 평가 점수를 수집하며, 평가점수에 의한 점수 편향을 방지하기 위해 평가 점수는 Chat Log에 기록되지 않는다. 이를 통해 권위 역할과의 상호작용이 General Public 에이전트의 평가 판단에 미치는 영향이 대화 진행에 따라 어떻게 변화하는지를 관찰할 수 있다.

\begin{table*}
\centering
\small
\begin{tabular}{l@{\;}l|c@{\;\;}c@{\;\;}c@{\;\;}c|c@{\;\;}c@{\;\;}c@{\;\;}c|c@{\;\;}c@{\;\;}c@{\;\;}c|c@{\;\;}c@{\;\;}c@{\;\;}c}
\toprule
& &\multicolumn{8}{c|}{\textbf{GPT-4o}} & \multicolumn{8}{c}{\textbf{Deepseek R1}}\\
& &\multicolumn{4}{c|}{\textbf{FairEval}} & \multicolumn{4}{c|}{\textbf{Topical-Chat}} & \multicolumn{4}{c|}{\textbf{FairEval}} & \multicolumn{4}{c}{\textbf{Topical-Chat}} \\
&  &$t_1$& $t_4$ & $t_8$ & $t_{12}$ &$t_1$& $t_4$ & $t_8$ & $t_{12}$ &$t_1$& $t_4$ & $t_8$ & $t_{12}$ &$t_1$& $t_4$ & $t_8$ & $t_{12}$ \\

\midrule

Judge & $A_t$& 86.3 & 85.0 & 83.8 & 85.0 & 58.3 & 65.0 & 60.0 & 56.7 & 85.0 & 93.8 & 95.0 & 95.0 & 83.3 & 96.7 & 96.7 & 96.7\\

& $\kappa_t$& 0.71 & 0.68 & 0.65 & 0.68 & 0.37 & 0.47 & 0.40 & 0.35 & 0.72 & 0.87 & 0.90 & 0.90 & 0.72 & 0.94 & 0.94 & 0.94\\
\cmidrule{3-18}  
Foreman & $A_t$ & 66.3 & 70.0 & 73.8 & 73.8 & 46.7 & 61.7 & 66.7 & 58.3 & 80.0 & 90.0 & 95.0 & 95.0 & 90.0 & 100 & 96.7 & 98.3 \\ 
& $\kappa_t$ & 0.43 & 0.47 & 0.53 & 0.54 &  0.19 & 0.42 & 0.49 & 0.37 &  0.64 & 0.81 &  0.90 &  0.90 & 0.81 & 1.00 & 0.93 & 0.96 \\

\cmidrule{3-18}  
Management & $A_t$ & 68.8 & 71.3 & 70.0 & 71.3 & 50.0 & 58.3 & 60.0 & 56.7 & 78.1 & 90.6 & 90.6 & 93.8 & 90.0 & 98.3 & 98.3 & 98.3 \\
&  $\kappa_t$ & 0.45 & 0.47 & 0.44& 0.47 & 0.19 & 0.39 & 0.41& 0.36 &  0.61 & 0.83 &  0.82 &  0.88 & 0.82 & 0.97 & 0.97 & 0.97 \\

\midrule

Supervisor & $A_t$ & 66.3 & 76.3 & 75.0 & 73.8 & 50.0 & 68.3 & 65.0 & 70.0  & 82.5 & 93.8 & 95.0 & 95.0 & 91.7 & 100 & 96.7 & 98.3  \\
&  $\kappa_t$ & 0.38 & 0.54 & 0.52 & 0.50 &  0.27 & 0.49& 0.44 &  0.52 & 0.68 & 0.88 &  0.90 &  0.90 & 0.83 & 1.00 & 0.93 & 0.97 \\
\cmidrule{3-18}  

Leader& $A_t$ & 62.5 & 68.8 & 68.8 & 70.0 & 46.7 & 65.0 & 66.7 & 63.3 & 81.3 & 96.3 & 96.3 & 93.8 & 85.0 & 98.3 & 98.3 & 100\\
&  $\kappa_t$ & 0.37 & 0.46 & 0.46 & 0.49 &  0.20 & 0.47 & 0.49 &  0.44 &  0.66 & 0.92 &  0.92 &  0.87 & 0.73 & 0.97 & 0.97 & 1.00 \\

\cmidrule{3-18}  
Mentor& $A_t$ & 62.5 & 71.3 & 72.5 & 68.8 & 38.3& 56.7 & 56.7 & 61.7 & 81.3 & 90.0 & 91.3 & 91.3 & 90.0 & 96.7 & 96.7 & 95.0\\

& $\kappa_t$ & 0.38 & 0.50 & 0.53 &  0.47 &  0.10 & 0.33 &  0.33 &  0.40 &  0.67 & 0.80 &  0.83 &  0.83 & 0.81 & 0.93 & 0.93 & 0.90 \\
\midrule

Specialist& $A_t$ & 63.8 & 71.3 & 71.3 & 71.3 & 46.7 & 58.3 & 56.7 & 56.7 & 81.3 & 96.3 & 95.0 & 95.0 & 93.3 & 100 & 100 & 100  \\
&  $\kappa_t$ & 0.35 & 0.47 & 0.46 &  0.48 &  0.18 & 0.40 &  0.37 &  0.37 &  0.65 & 0.92 &  0.89 &  0.89 & 0.87 & 1.00 & 1.00 & 1.00 \\
\cmidrule{3-18}  
Expert& $A_t$ & 70.0 & 76.3 & 75.0 & 76.3 & 50.0 & 63.3 & 65.0 & 63.3 & 82.5 & 92.5 & 95.0 & 95.0 & 91.7 & 100 & 100 & 100 \\
& $\kappa_t$ & 0.49 & 0.58 & 0.56 &  0.58 &  0.24 & 0.45 &  0.47 &  0.45 &  0.67 & 0.84 &  0.90 &  0.90 & 0.84 & 1.00 & 1.00 & 1.00 \\

\cmidrule{3-18}  
Attorney & $A_t$ & 60.0 & 70.0 & 70.0 & 68.8 & 38.3 & 56.7 & 55.0 & 55.0 & 85.0 & 91.3 & 93.8 & 91.3 & 90.0 & 100 & 100 & 100  \\
& $\kappa_t$ & 0.32 & 0.48 & 0.48 &  0.46 &  0.01 & 0.40 &  0.37 &  0.38 &  0.71 & 0.82 &  0.87 &  0.82 & 0.81 & 1.00 & 1.00 & 1.00 \\

\bottomrule

\end{tabular}
\caption{Experimental result of Free-form analysis. For each time step $t_i$ ($i=1,4,8,12$), each row present agreement ($A_t$) and Cohen's Kappa ($\kappa_t$) values between General Public and authority roles at $t_i$.}
\label{tab:rq1_main_result}

\end{table*}
\subsection{Evaluation}

% We measured authority bias through conversations between General Public agents and authoritative role agents following ChatEval's Single Agent evaluation methodology. 
% The experiment consisted of 12-turn conversations where the General Public agent initiated the dialogue in turn 1, followed by the authoritative role agent's response in turn 2. To measure authority bias, we first measured initial opinion of General Public and authority role, by collecting decision at turns 1 (for General Public) and 2 (for authoritative role). As the conversation continued until 12 turns, we further measured the decision of General Public at turns 4, 8, and 12, to identify changes. 
% To measure authority bias, General Public agents' evaluation scores were collected at turns 1, 4, 8, and 12, while authoritative role agents' scores were extracted at turn 2 when they first presented their opinions. 

To measure authority bias, we defined response selection agreement $A_t$ and Cohen's Kappa coefficient $\kappa_t$ between two agents, for each turn $t=1,4, 8$ and 12:
\begin{eqnarray*}
A_t &=& \text{Agreement}(S_{GP}^{(t)}, S_{Auth}^{(2)})\\ %\\, \quad t \in \{1, 4, 8, 12\}\\
\kappa_t &=& \text{Cohen's Kappa}(S_{GP}^{(t)}, S_{Auth}^{(2)})%, \quad t \in \{1, 4, 8, 12\}
\end{eqnarray*}
% where $S_{GP}^{(t)}$ represents the General Public agent's final choice at turn $t$, and $S_{Auth}^{(2)}$ represents the authoritative role agent's final choice at turn 2.

Specifically, using the authority agent's decision at turn 2 as a fixed reference point, we measured the degree of agreement with the General Public's decision at turns 1, 4, 8, and 12; thereby, we could track changes in authority bias as conversations progressed. Note that we can identify a increasing conformity to authority judgments when $A_t$ and $\kappa_t$ $(t=4,8,12)$ is higher than $A_1$ and $\kappa_1$, respectively.
% Note that higher $A_t$ and $\kappa_t$ values in subsequent turns ($t=4, 8, 12$) indicate increasing conformity to authority judgments as conversations progress. 
So, we operationally define this phenomenon of increasing agreement levels from $A_1$ to subsequent turns as authority bias.

%Stat에서 통계검정을 할 수 없는 이유 여기에 끝낼것
%\color{blue}
%For statistical analysis, we considered Bowker's test to assess symmetry in contingency tables. However, since our primary interest was in observing the concentration of values along the main diagonal of contingency matrices rather than testing symmetry, we focused on calculating Cohen's Kappa coefficient to measure the degree of agreement between agents across different time points.
%\color{black}

\subsection{Selected Models}
After conducting preliminary experiments with various state-of-the-art LLMs, we selected models capable of maintaining fluent and consistent dialogue in extended 12-turn discussions. 
Though we initially tested five models including QwQ \citep{qwq32b}, Gemini 2.5 Pro \citep{comanici2025gemini} and Mixtral \citep{jiang2024mixtral}, we found that only DeepSeek R1 \citep{guo2025deepseek} and GPT-4o \citep{hurst2024gpt} successfully finished all tasks without generating repeated responses.
Consequently, we utilized DeepSeek R1 and GPT-4o, which demonstrated fluent conversational abilities. All experiments were conducted with temperature set to 0 for reproducibility, and model calls were made through the OpenRouter API \citep{openrouterapi2025}.
%After conducting preliminary experiments with various state-of-the-art LLMs, we selected models capable of maintaining fluent and consistent dialogue in extended discussions 12 turns. Consequently, we utilized DeepSeek R1 \citep{guo2025deepseek}, GPT-4o \citep{hurst2024gpt}, which demonstrated fluent conversational abilities, in our experiments.
%본 연구에서는 다양한 최신 LLM들을 대상으로 예비 실험을 수행한 결과, 12턴까지의 장기간 토론에서 유창하고 일관된 대화를 유지할 수 있는 모델들을 선정하였다. 결과적으로 대화를 유창하게 한 DeepSeek R1 \citep{}, GPT-4o \citep{}, LLaMA4 Maverick \citep{}을 실험에 활용하였다. 

\subsection{Result and Discussion}
Referring to Table \ref{tab:rq1_main_result}, General Public follow the opinions of authority roles in most results. 
Across all experimental conditions, the agreement in $A_1$ was measured lower than subsequent turns ($A_4$, $A_8$, $A_{12}$), clearly confirming the presence of authority influence.
Particularly, differences between models and roles were prominently observed, while differences between turn numbers were not observed.
%Table 1를 참조하면, 대부분의 결과에서 General Public이 Authority의 의견을 따라가는 것을 확인할 수 있다. 특히, Model간 차이, Role간 차이가 두드러지게 나타났으며, Turn수의 차이는 무의미한 것을 확인하였다.

\begin{table}[t]
\centering
\small
\begin{tabular}{@{}l@{\;}|c@{\;\;}c|c@{\;\;}c@{}}
\toprule
& \multicolumn{2}{c|}{\textbf{GPT-4o}}  & \multicolumn{2}{c}{\textbf{Deepseek R1}}\\
& FairEval & TopicalChat &FairEval & TopicalChat \\
\midrule
Judege & 0.12 & 0.37 & 0.04 & 0.04\\
Foreman & 0.25 & 0.35 & 0.04 & 0.00\\
Management & 0.29 & 0.45 & 0.06 & 0.02 \\

\midrule

Supervisor& 0.24 & 0.27 & 0.04 & 0.00\\
Leader & 0.29 & 0.34 & 0.00 & 0.00\\
Mentor & 0.29 & 0.30 & 0.05 & 0.00\\

\midrule
Specialist & 0.28 & 0.40 & 0.03 & 0.00\\
Expert & 0.24 & 0.35 & 0.03 & 0.00\\
Attoney & 0.30 & 0.45 & 0.00 & 0.00\\

\bottomrule

\end{tabular}
\caption{Neutral response selection rates by authoritative roles at $A_2$ across models and datasets.}
\label{tab:rq1_neutral}
\end{table}
\paragraph{Model}
In the model comparative analysis, DeepSeek R1 showed significantly higher agreement levels than GPT-4o. 
For example, DeepSeek R1 showed $A_t$ of 100\% at least once in 6 out of 9 roles on Topical-Chat dataset, indicating greater susceptibility to authority influence compared to GPT-4o.
Additionally, DeepSeek R1's initial agreement ($A_1$) was higher than GPT-4o's, suggesting differences in the models' inherent sensitivity to authority. Particularly for GPT-4o, it was difficult to definitively conclude that authority influence occurred for roles with Legitimate Power (Judge, Foreman, Management). Examining the agreement change patterns for Legitimate Power roles, 5 out of 6 cases showed phenomena where agreement partially decreased as turns increased. When compared with DeepSeek R1 under identical conditions, GPT-4o consistently showed lower agreement levels across all turns. To identify the causes of these inter-model differences, we analyzed actual response distributions and found that GPT-4o had a tendency to excessively select neutral opinions, as confirmed in Table \ref{tab:rq1_neutral}: GPT-4o selected neutral options in 12\%-45\% cases, and Deepseek R1 selected them in less than 6\%. So, we suspect that the neutral responses might affect the outcome of this model comparative analysis.

\paragraph{Neutral Option} To further investigate this phenomenon, we separated GPT-4o's responses into two groups: cases where authoritative roles selected neutral options and cases where they selected non-neutral options, then measured authority bias for each group. Due to the space limitation, we describe some significant results here, to help readers understand the phenomenon. Appendix \ref{app:exp1_neutral} presents detailed results for each case.

In the group where authoritative roles selected only neutral options, the average differences between $A_1$ and $A_4$ for each authority type were: Legitimate group 36\%, Referent group 32.6\%, and Expert group 31.1\%. All roles showed notably decreased $A_t$ values in turns 4, 8, and 12 compared to turn 1. This confirms that General Public agents with GPT-4o are not actually influenced by authority when Authority roles makes neutral selections. 

Conversely, in the group excluding neutral options, the average differences between $A_1$ and $A_4$ for each authority type showed increases: Legitimate group 28.7\%, Referent group 32.3\%, and Expert group 35.9\%. Most roles demonstrated sharp increases in $A_t$ and $\kappa_t$ values in turns 4, 8, and 12 compared to turn 1, confirming that General Public agents with GPT-4o follow Authority's opinions when Authority roles took definite positions.

Overall, the difference between DeepSeek R1 and GPT-4o can be attributed to GPT-4o's higher rate of neutral responses. When authoritative roles provide neutral responses, General Public agents do not follow those opinions, representing an exceptional phenomenon where authority influence is negated. When agreement was remeasured excluding neutral options, GPT-4o's indicators were adjusted to levels similar to DeepSeek R1.

\paragraph{Role}
Through the results in Table \ref{tab:rq1_main_result}, we can observe patterns where authority influence varies according to roles. Analysis results show that role groups with Legitimate Power (Judge, Foreman, Management) exhibited relatively lower influence compared to other authority groups. Specifically, in GPT-4o on Faireval dataset, the average difference between $A_1$ and $A_4$ for the Legitimate Power group was 2.8\%, while the Referent Power group (Supervisor, Leader, Mentor) showed 7.5\%, and the Expert Power group (Specialist, Expert, Attorney) showed 7.2\% difference. This pattern appears consistently in DeepSeek R1 as well, and similar tendencies can be confirmed in the Topical-Chat dataset. Particularly, roles with Expert Power (Specialist, Expert, Attorney) were observed to reach complete agreement in all three roles under DeepSeek R1's Topical-Chat conditions.

To identify the causes of these differences between authority groups, we reviewed psychological literature and confirmed the characteristics of each authority type. According to \citet{carson1993social}, Legitimate Power is authority based on formal position or social status, Referent Power is authority derived from personal appeal or trust, and Expert Power is authority based on specialized knowledge or technical competence. Considering that this study dealt with evaluation tasks, evaluation situations have characteristics where accurate judgment and reliability are emphasized. Therefore, there is a possibility that Expert Power based on specialized knowledge or Referent Power based on trust may be perceived as more persuasive in evaluation contexts than Legitimate Power that relies on simple formal position. However, while we can consider the possibility that the patterns observed in this study may be related to the inherent differences between these authority types, further research is needed to draw clear conclusions.

When examining the results after excluding neutral options, these role-based differences become even more pronounced. Due to the space limitation, we only describe some significant results here. For instance, the Judge role in Faireval showed no change or even a 1.4\% decrease between turn 1 and subsequent turns 4, 8, and 12. Similarly, other Legitimate Group roles including Foreman and Management exhibited minimal agreement changes compared to other role groups. Conversely, Expert Group roles demonstrated the largest change magnitude among all groups. From a temporal perspective, the average changes after turn 4 were: Legitimate group 2.5\%, Referent group 2.8\%, and Expert group 3\%, maintaining consistently similar values throughout subsequent turns.

\begin{table*}[ht]
\centering
\small
\begin{tabular}{@{}l@{\;}l@{\;}|c@{\;\;\;}c@{\;\;\;}c@{\;\;\;}c|c@{\;\;\;}c@{\;\;\;}c@{\;\;\;}c|c@{\;\;\;}c@{\;\;\;}c@{\;\;\;}c|c@{\;\;\;}c@{\;\;\;}c@{\;\;\;}c@{}}
\toprule
& &\multicolumn{8}{c|}{\textbf{GPT-4o}} & \multicolumn{8}{c}{\textbf{Deepseek R1}}\\
& &\multicolumn{4}{c|}{\textbf{FairEval}} & \multicolumn{4}{c|}{\textbf{Topical-Chat}} & \multicolumn{4}{c|}{\textbf{FairEval}} & \multicolumn{4}{c}{\textbf{Topical-Chat}} \\
&  &$t_1$& $t_4$ & $t_8$ & $t_{12}$ &$t_1$& $t_4$ & $t_8$ & $t_{12}$ &$t_1$& $t_4$ & $t_8$ & $t_{12}$ &$t_1$& $t_4$ & $t_8$ & $t_{12}$ \\
\midrule
General & $A_t$ & 91.3 & 98.8 & 97.5 & 97.5&71.7&75.0&73.3 & 73.3&93.8&98.8&98.8&98.7&86.7&91.7&90.0&90.0\\
Public & $\kappa_t$ & 0.81 & 0.97 & 0.94 & 0.94 & 0.58 &  0.61 &  0.58 &  0.58 & 0.89 & 0.98 & 0.98 & 0.98&0.75 &  0.84& 0.81&0.81\\
\midrule
Judge & $A_t$ & 53.8 & 62.5 & 61.3 & 60.0 & 63.3 & 61.7& 63.3 & 66.7& 87.5 & 98.8 & 98.8 & 98.8 & 85.0 & 96.7 & 96.7 & 98.3\\
& $\kappa_t$ & 0.27 & 0.41 & 0.39 & 0.38 & 0.45 & 0.44 & 0.46 & 0.51 & 0.77 &0.97 & 0.97 & 0.97 & 0.73 & 0.93 & 0.93 & 0.97\\
\cmidrule{3-18} 
Foreman & $A_t$ & 68.8 & 70.0 & 70.0 & 70.0 & 53.3 & 60.0& 60.0 & 63.3& 87.5 & 98.8 & 98.8 & 100.0 & 88.3 & 98.3 & 98.3 & 98.3\\
& $\kappa_t$ & 0.47 & 0.48 & 0.48 & 0.48 & 0.28 & 0.42 & 0.42 & 0.47  & 0.77 & 0.97 & 0.97 & 1.00 & 0.79 & 0.97 & 0.97 & 0.97  \\
\cmidrule{3-18} 
Management & $A_t$ & 55.0  & 48.3 & 51.7 & 48.3 & 70.0 & 75.0 & 73.8 & 70.0& 86.3 & 96.3 & 97.5 & 97.5 & 90.0 & 96.7 & 98.3 & 98.3\\
& $\kappa_t$ & 0.36 & 0.47 & 0.43 & 0.45 & 0.29 & 0.30 & 0.34 & 0.30 & 0.75 & 0.93 & 0.95 & 0.95 & 0.82 & 0.93 & 0.97 & 0.97\\

\midrule
Supervisor & $A_t$ & 65.0 & 68.8& 70.0 & 71.3 & 46.7 & 58.3 & 63.3 & 61.7& 86.3 & 97.5 & 98.8 & 97.5 & 88.3 & 98.3 & 96.7 & 96.7\\
& $\kappa_t$ & 0.39 & 0.45 & 0.47 & 0.49 & 0.20 & 0.38 & 0.45 & 0.43 & 0.75 & 0.95 & 0.97 & 0.95 & 0.79 & 0.97 & 0.94 & 0.94\\
\cmidrule{3-18} 
Leader & $A_t$ & 70.0 & 75.0 & 73.8 & 70.0 & 48.3 & 66.7 & 66.7 & 66.7& 81.3 & 100.0 & 98.8 & 98.8 & 86.7 & 96.7 & 98.3 & 98.3\\
& $\kappa_t$ & 0.50 & 0.58 & 0.55 & 0.49 & 0.22 & 0.50 & 0.50 & 0.50 & 0.67 & 1.00 & 0.97 & 0.97 & 0.76 & 0.93 & 0.97 & 0.97\\
\cmidrule{3-18} 
Mentor & $A_t$ & 73.8 & 76.3 & 75.0 & 75.0 & 53.3 & 71.7 & 66.7 & 68.3& 81.3 & 97.5 & 96.3 & 96.3 & 90.0 & 96.7 & 96.7 & 98.3\\
& $\kappa_t$ & 0.54 & 0.58 & 0.56 & 0.32 & 0.32 & 0.55 & 0.48 & 0.50 & 0.67 & 0.95 & 0.93 & 0.92 & 0.81 & 0.93 & 0.93 & 0.97\\

\midrule
Specialist & $A_t$ & 61.3 & 65.0 & 65.0 & 66.3 & 55.0 & 63.3 & 66.7 & 63.3& 80.0 & 97.5 & 96.3 & 96.3 & 85.0 & 96.7 & 98.3 & 98.3\\
& $\kappa_t$ & 0.37 & 0.42 & 0.42 & 0.45 & 0.32 & 0.45 & 0.51 & 0.46 & 0.64 & 0.95 & 0.93 & 0.92 & 0.74 & 0.93 & 0.97 & 0.97\\
\cmidrule{3-18} 
Expert & $A_t$ & 65.0 & 70.0 & 70.0 & 70.0 & 56.7 & 51.7 & 55.0 & 53.3& 85.0 & 98.8 & 98.8 & 100.0 & 86.7 & 95.0 & 98.3 & 98.3\\

& $\kappa_t$ & 0.43 & 0.50 & 0.50 & 0.50 & 0.32 & 0.31 & 0.37 & 0.34 & 0.73 & 0.97 & 0.97 & 1.00 & 0.76 & 0.90 & 0.97 & 0.97\\
\cmidrule{3-18} 
Attorney & $A_t$ & 66.3 & 77.5 & 77.5 & 76.3 & 46.7 & 73.3 & 75.0 & 75.0& 82.5 & 98.8 & 97.5 & 97.5 & 91.7 & 98.3 & 96.7 & 96.7\\
& $\kappa_t$ & 0.40 & 0.59 & 0.59 & 0.56 & 0.24 & 0.57 & 0.60 & 0.60 & 0.70 & 0.97 & 0.95 & 0.95 & 0.84 & 0.97 & 0.93 & 0.93\\
\bottomrule
\end{tabular}
\caption{Experimental result of Content-controlled analysis. For each time step $t_i$ ($i=1,4,8,12$), each row present agreement ($A_t$) and Cohen's Kappa ($\kappa_t$) values between General Public and authority roles at $t_i$.}
\label{tab:rq2_main}
\end{table*}

\paragraph{Turn}
When observing changes according to turns, we analyzed the change patterns of $A_4$, $A_8$ and $A_{12}$, excluding $A_1$ which was not influenced by authoritative roles. As a result, we \textbf{did not observe typical authority bias} patterns where agreement continuously increased with increasing turn numbers. Instead, the following two patterns frequently appeared: 1) $A_4 = A_8 = A_{12}$ and 2) $A_4<A_8=A_{12}$, which occurred in 5 and 10 cases for DeepSeek R1, and 2 cases each for GPT-4o. The emergence of these irregular patterns suggests the possibility that they were influenced by \textit{conversational content} rather than pure authority influence. In free-form conversations, new information or arguments are presented at each turn, and these content-related factors are presumed to have affected changes in $A_t$. These findings indicate the necessity for a controlled experiment to isolate the pure effect of authoritative role from content.

\section{Experiment 2 : Controlling Content}
Experiment 2 was designed to extend the findings of Experiment 1 and identify where authority bias specifically manifests. To distinguish whether the observed authority bias stems from roles themselves or conversational content differences, we conducted a controlled experiment that regulates conversational content. Through this approach, we aim to more accurately identify the causes of authority bias and provide in-depth analysis of the sources of authority bias arising from roles in MAS.
%Experiment 2는 Experiment 1의 연구 결과를 확장하여 권위편향이 구체적으로 어디서 나타나는지를 확인하기 위해 설계하였다. Free-form 대화에서 관찰된 권위편향이 역할 자체에서 비롯되는 것인지, 아니면 대화 내용의 차이에서 발생하는 것인지를 구분하기 위해 대화 내용을 통제한 실험을 수행한다. 이를 통해 권위편향의 원인을 보다 정확히 파악하고, Multi-Agent 시스템에서 Role에서 발생되는 권위 편향의 원인을 심층적으로 분석하고자 한다. 

\subsection{Experiment Procedure}
The experimental procedure was identical to Experiment 1 except how we used the authoritative roles. In this experiment, we utilized conversation datasets between two General Public agents, instead of using conversation between General Public and authoritative roles. For Public-Authoritative conversations, we only altered the second agent's role from General Public to one of the nine authoritative roles while maintaining the contents of Public-Public conversation unchanged. This design allows us to isolate and observe the pure effect of role labeling on decisions while eliminating the influence of conversational content.
%
% The experimental procedure was identical to Experiment 1 except for changing the role of the second General Public agent to an authoritative role in the existing General Public conversation logs. 
%
The models used, datasets, evaluation turns, and evaluation score collection methods were all applied identically to Experiment 1. The authority bias measurement methodology also utilized the same formulas from Experiment 1 to ensure consistency of results.
%본 실험에서는 두명의 General Public 간의 대화 데이터셋을 활용한다. 수집된 대화 내용은 그대로 유지하되, 한쪽 에이전트의 역할만을 9가지 권위 역할 중 하나로 변경하여 권위편향을 측정한다. 이러한 설계를 통해 대화 내용의 영향을 배제하고 순수하게 역할 라벨링이 평가 점수에 미치는 영향만을 분리하여 관찰할 수 있다. 실험을 위해서 기존의 General Public간의 대화로그에서 General Public 2의 Role을 변경한 것을 제외한 모든 실험 절차는 Experiment 1과 동일하다. 사용된 모델, 데이터셋, 평가 시점 그리고 평가 점수 수집 방법 모두 Experiment 1과 동일하게 적용된다. 권위편향 측정 방법론 또한 Experiment 1에서 사용한 수식과 통계 분석 방법을 그대로 활용하여 결과의 일관성을 보장한다.

\subsection{Result and Discussion}
As shown in Table \ref{tab:rq2_main}, we observed conversations between General Public agents where only one agent's role name was changed to an Authoritative role. As a result, two major findings emerged. First, there were significant differences between Public-Public conversations and Public-Authoritative conversations. Second, Similar patterns to experiment 1 results appeared despite controlling for content.
%General Public 간의 대화에서 한쪽 role의 이름만 Authority로 변경한 상황을 관찰한 결과, 두가지 주요 finding이 있었다. 1) General Public간의 대화에서는 General Public과 Authoritative role간의 대화와 큰 차이가 있었다는 점. 2) Content를 통제하였음에도 불구하고 RQ1의 결과와 유사한 패턴이 나타난 점.

\paragraph{Overall} 
Upon comparing the results between Public-Public conversation and Public-Authoritative conversation, notable differences were confirmed in $A_1$ and $\kappa_1$ values. The $A_1$ values for non-authoritative General Public agents were significantly higher than those in cases with authoritative roles. Based on these observations, it can be interpreted that while the second General Public was influenced by opinion of the first General Public, Authoritative roles may not have been influenced by General Public responses, which is similar result to Experiment 1. Analyzing these patterns, it could be interpreted that rather than General Public agents being influenced by Authority, Authoritative roles tend to show less conformity to General Public opinions.
%우선 General Public 간의 결과와 General Public-Authority 역할 간의 결과를 비교 분석한 결과, $A_1$과 $\kappa_1$에서 현저한 차이가 확인되었다. 권위를 갖지 않은 General Public의 $A_1$ 값이 권위 역할을 가진 경우의 $A_1$보다 현저히 높게 나타났다. 이러한 점을 관찰했을 때 General Public2가 General Public1의 응답에 영향을 받았지만, Authoritative role은 General Public의 응답에 영향을 받지 않을 것으로 생각할 수 있다. 이러한 패턴을 분석하면, 일반적으로 General Public이 Authority의 영향을 받는다기보다는 Authority 역할이 General Public의 의견을 따라가지 않는 경향이 있다고 해석할 수 있다.

\paragraph{Neutral Option} 
These results became more pronounced when removing GPT-4o's neutral option responses, as mentioned in experiment 1. Due to the space limitation, we describe some significant results here, and detailed results are presented in Appendix \ref{app:exp2_neutral}.

When measuring authority bias in GPT-4o's neutral option group versus non-neutral options, the neutral group showed decreased $A_t$ values in turns 4, 8, and 12 compared to turn 1 across most roles. This suggests a phenomenon similar to experiment 1, where GPT-4o may not actually be influenced by authority when making neutral choices. 

In the group excluding neutral options, sharp increases in both $A_t$ values and $\kappa_t$ values were observed in turns 4, 8, and 12 compared to turn 1. This could suggest that General Public agents follow Authority opinions. Comprehensively considering this phenomena on non-neutral option, GPT-4o becomes showing similar patterns to DeepSeek R1. Detailed results are in Appendix~\ref{app:exp1_neutral} and \ref{app:exp2_neutral}.
%이러한 결과는 RQ1에서 언급했던 GPT-4o의 중립 옵션에 대한 응답을 제거하고 관찰한 결과에서 더 두드러지게 나타났다. gpt-4o의 RQ2 뉴트럴 옵션그룹과 뉴트럴 옵션을 제외한그룹에서 권위편향을 측정해본 결과, 뉴트럴 선택지만 존재하는 그룹에서는 대부분의 role에서 agreement값이 turn1에 비해 turn4,8,12에서 줄어드는 것을 확인하였다. 이러한 점은 gpt-4o가 중립선택을 했을 경우에 실제로 권위의 영향을 받지 않는다는 RQ1과 유사한 현상이 관찰되었다. 뉴트럴 옵션을 제외한 그룹을 보면, Role이 turn1에 비해 turn 4,8,12의 agreement 값과 kappa값이 급격히 상승하는 것을 확인할 수 있다. 이로인해, General Public이 Authority의 의견을 따라가는 현상을 확인할 수 있다. 이러한 현상을 종합적으로 보면, GPT-4o와 Deepseek R1의 결과가 유사하게 나타나였다. 자세한 결과는 Supplymentary Matarial을 확인하라. 

\paragraph{Role}
Role effects also demonstrate similar patterns to experiment 1. Despite controlling for content influence, Legitimate roles (Judge, Foreman, Management) showed weaker impact compared to other authority types. Analysis of Table \ref{tab:rq2_main} reveals that both GPT-4o and DeepSeek R1 exhibited relatively lower increases in $A_t$ for the Legitimate Power group compared to other authority types. Notably, Expert Power roles in DeepSeek R1 maintained high agreement levels above 95\% even when controlling contents, reconfirming strong expertise-based authority in evaluation tasks. When compared to Public-Public conversations, the inclusion of Authoritative roles consistently resulted in markedly lower $A_1$ values across all authority types. This suggests that authority bias stems from the authoritative characteristics of roles themselves rather than conversational content. Particularly, the pattern observed in human social psychology where Expert and Referent Power exert stronger influence than Legitimate Power in evaluation contexts \citep{carson1993social} is reproduced in LLMs.
%Role의 영향도 마찬가지로 RQ1과 유사한 패턴이 관찰된다. Content의 영향을 통제하였음에도 불구하고 Legitimate의 Role (Judge, Foreman, Management)가 영향이 적은 것을 확인할 수 있다. Table 3의 결과를 분석해보면, GPT-4o와 DeepSeek R1 모두에서 Legitimate Power 그룹이 다른 권위 유형에 비해 상대적으로 낮은 agreement 증가를 보였다. 특히 DeepSeek R1에서 Expert Power 역할들이 Content Control 조건에서도 95\% 이상의 높은 agreement를 유지한 것은 평가 과제에서 전문성 기반 권위의 강력한 영향력을 재확인한다. General Public 간 대화와 비교했을 때 Authority 역할이 포함된 경우 $A_1$ 값이 현저히 낮아지는 현상이 모든 권위 유형에서 일관되게 나타났다. 이는 권위편향이 대화 내용이 아닌 역할 자체의 권위적 특성에서 비롯됨을 시사한다. 특히 평가 맥락에서는 Legitimate Power보다 Expert Power와 Referent Power가 더 강한 영향력을 갖는다는 인간 사회심리학의 패턴 \citep{carson1993social}이 LLM에서도 재현되고 있음을 확인할 수 있다.

\paragraph{Turn} 
The temporal dynamics of authority bias in experiment 2 mirror the patterns observed in experiment 1. When examining turn-by-turn changes, we did not observe typical patterns of authority bias, i.e., continuously increasing agreement. Instead, two patterns emerged as in experiment 1: 1) $A_4 = A_8 = A_{12}$ and 2) $A_4 < A_8 = A_{12}$, appearing in 8 cases for DeepSeek R1 and 3 cases for GPT-4o. Critically, when considering $A_1$, authority-absent conversations maintained high initial agreement throughout subsequent turns, while authority-present conversations showed persistent low agreement. This suggests that in authority-absent dialogues, initial content continues to influence decision making through the conformity of General Public agents. Whereas, in authority-present dialogues, the independent nature of authoritative roles from General Public opinions weakens the influence of turn 1 while amplifying the impact of turn 2. This pattern remained consistent even when controlling contents in experiment 2, confirming that temporal dynamics of authority bias may stem from structural power relationships rather than conversational content.
%Turn의 영향도 RQ1과 마찬가지로 유사한 패턴이 관찰된다. turn에 따른 변화를 관찰해보았을때, typical authority bias pattern은 발견되지 않으나 1) $A_4$ = $A_8$  = $A_12$  2) $A_4$ $<$ $A_8$, $A_8$ = $A_{12}$, 이것들은 Deepseek R1에서는 8개, GPT-4o에서는 3개가 나타났다. $A_1$의 값을 함께 고려해보면, authority가 없는 역할의 경우 초기의 높은 agreement가 Turn 4 이후에도 안정적으로 유지되는 반면, authority가 있는 역할의 경우에는 낮은 agreement가 Turn 4 이후까지 지속되는 경향이 나타났다. General Public 2가 General Public의 의견을 따라가는 것을 고려할 때, authority가 없는 role이 포함된 대화에서는  Turn 1의 의견이 이후 평가자들에게 지속적으로 반영되는 것으로 해석할 수 있다.반면 authorative role이 General Public의 의견을 따라가는 것을 고려할 때, authority 역할이 포함된 대화에서는 Turn 1의 영향력은 상대적으로 약화되며 Turn 2의 응답이 보다 강하게 반영되는 것으로 해석할 수 있다. 

\paragraph{Overall Discussion} 
Results of experiment 2 provide important insights into the nature of authority bias in MAS. The most significant finding is that the mechanism of authority bias differs from previous assumptions. While existing study assumed that General Public agents are actively influenced by Authority opinions, our results revealed different pattern. Public-Public conversations show high initial agreement $A_1$, whereas conversations including Authoritative roles exhibit markedly lower $A_1$ values. This suggests that Authoritative roles tend to maintain their positions without being influenced by other opinions. Consequently, the observed authority bias can be interpreted not as General Public agents actively conforming to Authority, but as a phenomenon resulting from the combination of two tendencies: (1) Authority agent persists in maintaining its position, and (2) General Public agent is too flexible to agree with others' opinion. 

Also, the emergence of identical patterns to experiment 1 despite controlling for conversational content confirms that such authority bias stems from role labels themselves rather than content. This indicates that LLMs recognize the social authority structures inherent in roles and exhibit corresponding behavioral patterns. Moreover, neutral option analysis and temporal patterns demonstrate that this mechanism is established early and persists over time. Authoritative roles' position maintenance and General Public's gradual conformity are formed in the early stages of content and subsequently stabilize throughout the discussion.

These results present a new perspective that authority bias in MAS is based on independence and consistency of Authoritative roles rather than mutual influence, suggesting that future MAS design requires bias mitigation strategies that consider such asymmetric interaction patterns.
%Content Control 실험의 결과는 Multi-Agent 시스템에서 나타나는 권위편향의 본질에 대한 중요한 통찰을 제공한다. 가장 중요한 발견은 권위편향의 메커니즘이 기존에 가정했던 것과 다르다는 점이다. 기존 연구들은 General Public이 Authority의 의견에 능동적으로 영향을 받는다고 가정했으나, 본 연구 결과는 이와 다른 패턴을 보여준다. General Public 간의 대화에서는 높은 초기 agreement($A_1$)를 보이는 반면, Authority 역할이 포함된 경우 현저히 낮은 $A_1$ 값을 보인다. 이는 Authority 역할이 다른 의견에 영향을 받지 않고 자신의 입장을 고집하는 경향이 있음을 시사한다. 결과적으로 관찰되는 권위편향은 General Public이 Authority에게 적극적으로 동조하는 것이 아니라, Authority의 일관된 주장 유지와 General Public의 상대적 유연성이 결합되어 나타나는 현상으로 해석할 수 있다.

%둘째, 대화 내용을 통제했음에도 불구하고 RQ1과 동일한 패턴이 나타난 것은 이러한 권위편향이 논리적 설득력이나 대화 내용이 아닌 역할 라벨 자체에서 비롯됨을 확인한다. 이는 LLM들이 역할에 내재된 사회적 권위 구조를 인식하고 그에 따른 행동 패턴을 보인다는 것을 의미한다.

%셋째, 중립 옵션 분석과 시간적 패턴은 이러한 메커니즘이 초기에 확립되어 지속됨을 보여준다. Authority 역할의 입장 고수와 General Public의 점진적 동조가 대화 초기에 형성되고 이후 안정화되는 특성을 보인다. 이러한 결과들은 Multi-Agent 시스템에서 권위편향이 상호 영향보다는 Authority 역할의 독립성과 일관성에 기반한다는 새로운 관점을 제시하며, 향후 Multi-Agent 평가 시스템 설계에서 이러한 비대칭적 상호작용 패턴을 고려한 편향 완화 전략이 필요함을 시사한다.

\section{Conclusion}
Our study presents the first systematic analysis of role-based authority bias in Multi-agent systems. Through free-form and content-controlled experiments using ChatEval, we demonstrated that authority bias stems from inherent role characteristics rather than conversational content. Our findings reveal that referent and expert power roles exert stronger influence than legitimate power roles, mirroring human social psychology theory. Crucially, authority bias operates not through active conformity by general agents, but through a mechanism where authoritative roles maintain their positions while general agents demonstrate flexibility. These insights provide foundational knowledge for designing multi-agent frameworks where asymmetric interaction patterns significantly affect outcomes.%Through free-form and content-controlled experiments using the ChatEval framework, we demonstrated that authority bias stems from the inherent characteristics of roles rather than conversational content. Our findings reveal that referent and expert power roles exert stronger influence than legitimate power roles, mirroring patterns in human social psychology. Crucially, authority bias operates not through active conformity by general agents, but through a mechanism where authoritative roles consistently maintain their positions while general agents demonstrate flexibility in adjusting opinions. These insights provide foundational knowledge for designing multi-agent frameworks where asymmetric interaction patterns significantly affect outcomes.

\section*{Limitations}
This study systematically analyzes role-based authority bias in multi-agent evaluation systems, but two key limitations should be acknowledged. First, experiments were conducted using GPT-4o and DeepSeek R1, selected for their capability to maintain fluent dialogue across 12-turn conversations; however, this requirement constrained the diversity of models examined. Second, our ChatEval-based framework was designed to capture authority bias in evaluation tasks, yet multi-agent systems are deployed across diverse domains such as creative collaboration, technical problem-solving, and strategic planning. These domains may exhibit different authority bias patterns that our evaluation-focused design does not address.
\label{sec:reference_examples}

\bibliography{aaai2026}
\newpage
\appendix

\section*{The Use of Large Language Models}
We used AI assistance tools during the writing process of this manuscript. Specifically, we employed Grammarly for grammar checking, and GPT-5 for language polishing and improving clarity of expression. These tools were used for editorial purposes.

\section{Experiment Setup}
\label{sec:expsetup}
\subsection{Experimental Framework}
All experiments were conducted using the ChatEval \citep{chan2023chateval} framework, maintaining identical configurations to the original implementation. We preserved all system prompts, instructions, and dialogue structures from ChatEval, modifying only the role assignments to examine authority effects. Detailed implementation and prompts can be found in our code: [Under Review].

\subsection{Prompt Setup}
To isolate the effect of role-based authority, we kept all prompt components constant except for the role name and its co-reference in the role description. The following templates compares the original attempt and our approach. The first template indicates general public agent specified in the original ChatEval framework:
\begin{mdframed}

Name: \textbf{General Public}

Role description:\\
You are now \textbf{General Public}, one of the referees in this task.
You are interested in the story and looking for updates on the investigation.
Please think critically by yourself and note that it is your responsibility
to choose which response is better.
\end{mdframed}

And, the next template shows how we modified the basic template with authority roles.
\begin{mdframed}
Name: \textbf{[Authority Role]}

Role description:\\
You are now \textbf{[Authority Role]}, one of the referees in this task.
You are interested in the story and looking for updates on the investigation.
Please think critically by yourself and note that it is your responsibility
to choose which response is better.
\end{mdframed}

Where \textbf{[Authority Role]} is replaced with one of the nine authoritative roles: Judge, Foreman, Management, Supervisor, Leader, Mentor, Specialist, Expert, or Attorney.
      
\section{Proof of Authority Role Design}
In this section, we describe the detailed justification for our selection of nine authority roles inspired by \citet{french1959bases}.

\subsection{Legiminage Power}
\citet{french1959bases} defined legitimate power as the influence that emerges when \textit{P} accepts that \textit{O} has rightful authority to give directives and feels obligated to comply. This power stems from social and normative approval granted to positions, creating voluntary compliance because individuals perceive the authority as rightful.
%French와 Raven(1959)은 legitimate power을 “지시할 정당한 권리가 O에게 있다고 P가 받아들이며, 동시에 P가 그 지시에 따를 의무가 있다고 느낄 때 생기는 권력”으로 정의한다. 해당논문에서는 정당성이 행동을 유도하는 심리적 힘으로 기능하여, 직위가 명령할 권리와 개인이 따를 의무를 자연스럽게 만들어 낸다고 주장한다. 결국, 직책에 부여된 사회적·규범적 승인이 권력의 원천으로 작용하고 개인은 이를 ‘정당하다’고 느끼기에 자발적으로 순응하게되는 것이  legitimate power이다.  

\paragraph{Judge}
A judge represents a role with institutionally granted authority to make definitive judgments and render final decisions that others are expected to accept. This role embodies the power to evaluate situations and pronounce authoritative verdicts based on recognized legitimate right to make binding determinations. \citet{french1959bases} presented judge as an exemplary profession where legitimate power manifests through formal position-based authority.
%judge는 어떤 대상을 기준에 따라 심사하고 우열를 정하는 역할로써 공식 심사 권한을 갖고있다. French와 Raven(1959)에서는  판사는 벌금을 부과할 권리가 있는 직책이고 어떤 직책을 정당하다고 수용하는 것은 정당한 권력의 한 근거가 된다고 말하며 judge를 legitimate power가 나타나는 직업의 예시로 제시했다.  ㅇㅇ

\paragraph{Foreman}
A foreman represents a role with formally designated authority to direct and coordinate others' actions within structured hierarchies. This position embodies the power to assign tasks and ensure compliance through officially recognized right to make operational decisions. \citet{french1959bases} described foreman as demonstrating legitimate power through institutionally sanctioned supervisory authority.
%Foreman은 조직을 직접 지휘하고 감독하는 최일선 관리직으로써  조직 위계상 ‘반장’ 직위가 부여하는 공식적 통솔 권한을 지닌다.  French와 Raven(1959)에서는 작업반장은 업무를 배정해야 하는 정당성이 부여된 직책으로써 정당한 권력의 한 근거가 된다고 말하며 foreman를 legitimate power가 나타나는 직업의 예시로 제시했다. ㅇㅇ
\paragraph{Management}
Management represents roles with institutionally sanctioned authority to make strategic decisions and control organizational resources. This encompasses positions with formal power to set policies and direct organizational behavior through recognized executive decision-making rights. \citet{french1959bases} identified management as exemplifying legitimate power through institutional mandate to make authoritative choices.
%Management는  자원을 계확하고 통제하는 의사결정·조정 역할을 담당하는 직위이다. 일반적으로 중간관리자(supervisors, middle managers)부터 최고경영자(CEO·임원)에 이르는 “관리 기능을 수행하는 직책 전체” 를 포괄한다.  French와 Raven(1959)에서는 특정 결정은 경영진의 고유 권한이라 경영진이 legitimate power가 나타나는 직업이라고 주장했다. ㅇㅇ

\subsection{Reference Power}
\citet{french1959bases} defined referent power as influence that emerges when \textit{P} seeks to identify with \textit{O} and feels attraction and respect toward \textit{O}. This represents power arising from voluntary compliance driven by \textit{P}'s motivation to become like \textit{O}.

\paragraph{Supervisor}
A supervisor embodies a role that commands respect and voluntary compliance through demonstrating exemplary work practices and care for team development. This position represents the power to influence through being perceived as someone whose methods and approaches others want to emulate and \citet{haller2018power} identified supervisors as roles that team members respect and wish to model their work practices after, establishing supervisor as a position where referent power manifests.
%Supervisor는  생산·서비스·사무 등 다양한 현장에서 팀을 직접 관리하며, 팀원들에게 동기부여 등 작업의 전반적인 과정을 책임지는 최초 관리 계층(first-line management)이다. French & Raven의 다섯가지 권력체계를 계승한 논문인(Schneider et al. (2018) Examining the Relationship Between Leaders’ Power Use and Motivational Outlooks)에서는 Supervisor가 존경하고 업무방식을 본받고싶은 역할로 나타났다. 이로인해 supervisor을  referent power가 나타나는 직책이라고 할 수 있다. ㅇㅇ
 
\paragraph{Leader}
A leader represents a role that influences others through vision and inspiration, creating voluntary followership based on admiration for their character and direction. This position embodies the power to guide groups through personal magnetism and the ability to make others want to align themselves with the leader's mission. \citet{french1959bases} noted that referent power emerges when followers admire leaders and seek to identify with them as role models, positioning leader as an exemplar of referent power.
%Leader란 집단에서 방향을 제시하고, 구성원을 조정하거나 동기부여하여 목표 달성을 이끄는 역할이다. leader은 공식 직책이 없어도 영향력으로 집단을 이끄는 개인도 포함된다. (The Power of Good: A Leader’s Personal Power as a Mediator of the Ethical Leadership–Follower Outcomes Link.)에서는 referent power가 부하가 리더를 동경하여 자신을 동일시하고 역할 모델로 인식하는 데서 비롯된다고 언급하여 leader을 Referent Power의 예시로 들고있다.

\paragraph{Mentor}
A mentor represents a role that wields influence through being perceived as a wise guide whose experience and counsel others actively seek and value. This position embodies the power to shape development through being seen as someone worth emulating in both professional and personal growth. \citet{godshalk2000does} observed that mentees commonly attribute referent power to mentors, establishing mentor as a role where referent power naturally emerges.
%mentor는 경험과 지식을 바탕으로 맨티의 성장과 목표 달성을 돕는 조언자이자 롤모델이고, 공식적인 관계나 비공식 관계 모두 포함되는 직책이다. (Does Mentor-Protégé Agreement on Mentor Leadership Behavior Influence the Quality of a Mentoring Relationship?)에서는 멘티들이 멘토에게 준거 권력을 부여하는 경우가 흔하다고 말하며 mentor에게서 Referent Power가 나타난다고 본다.

\subsection{Expert Power}
\citet{french1959bases} defined expert power as influence that forms when someone is perceived to possess special knowledge or expertise. When \textit{P} believes \textit{O} has superior knowledge and credible expertise, \textit{P} voluntarily follows \textit{O}'s guidance, maintaining compliance even without rewards or punishments.
%(French, J. R. P. \& Raven, B. The Bases of Social Power.)에서는 expert power을 누군가가 특별한 지식이나 전문성을 지녔다고 인식될 때 형성되는 권력이라고 정의한다.  P는 O가 문제 판단에 있어서 자신보다 더 나은 지식를 갖고 있다고 믿고, O의 전문성이 신뢰도 있다고 평가되면, P는 O의 지시를 따르는 것이 최선의 선택이라고 판단하여 자발적으로 행동을 수정하므로, 보상·처벌이 없어도 순응이 비교적 안정적으로 유지된다고 서술한다.

\paragraph{Specialist}
A specialist represents a role that commands deference through possessing concentrated, domain-specific knowledge that others recognize as superior in particular areas. This position embodies the power to influence decisions through demonstrated mastery of specialized information and techniques. \citet{french1959bases} noted that expert power emerges specifically when others perceive someone as having special knowledge in defined domains, making the specialist role a direct manifestation of expertise-based influence within limited fields.

%Specialist는 조직 또는 직무 영역에서 한정된 분야에 깊이 있게 특화된 지식과 기술을 보유하여, 해당 영역의 문제를 효과적으로 수행하는 전문가적 직책이다. %(French, J. R. P. \& Raven, B. The Bases of Social Power.)에서는 expert power을 누군가가 특별한 지식이나 전문성을 지녔다고 인식될 때 형성되는 권력이라고 정의한다. specialist는 조직에서 특별한 지식이나 전문성을 지녔다고 인식받는다는 점에서 expert power가 발생한다고 볼 수 있다.

\paragraph{Expert}
An expert embodies a role with comprehensive mastery that others acknowledge as authoritative within specific domains. This position represents the power to shape opinions through demonstrated competence and superior analytical capability that provides persuasive decision-making resources. \citet{french1959bases} emphasized that expert power stems from recognized superior knowledge and judgment abilities across broader areas of expertise, positioning the expert role as the archetypal example of comprehensive expertise-based authority.

%Expert는 특정 분야에서 심도 있는 지식과 판단능력을 인정받아, 조직원들에게 설득력높은 의사결정 자원을 제공하는 최고 수준의 전문 직책이다. (French, J. R. P. \& Raven, B. The Bases of Social Power.)에서는 expert power을 누군가가 특별한 지식이나 전문성을 지녔다고 인식될 때 형성되는 권력이라고 정의한다. expert는 조직에서 특별한 지식이나 전문성을 지녔다고 인식받는다는 점에서 expert power가 발생한다고 볼 수 있다. 

\paragraph{Attorney}
An attorney represents a role with specialized analytical and argumentative capabilities that others recognize as essential for navigating complex evaluative processes. This position embodies the power to influence through systematic reasoning and structured analysis that others find compelling and trustworthy. \citet{french1959bases} specifically cited accepting legal counsel as a common example of expert influence in action, identifying the attorney role as a prime illustration of how specialized knowledge creates authoritative influence in decision-making contexts.
%Attorney은 법률 지식을 바탕으로 개인·기업·공공기관을 대신해 법률 자문, 계약과 문서 작성, 소송·중재 대리 등을 수행하는 전문직이다. (French와 Raven(1959)에서는 법률 문제에서 변호사의 조언을 받아들이는 것을 전문적 영향력의 흔한 예로 들며 attorney를 Expert Power를 나타내는 직업의 예시로 언급하였다. 이로써 attorney는 Expert Power을 잘 드러낼 수 있는 권위역할로 적합하다.
\\

\section{Neutral response on Experiment 1}
\label{app:exp1_neutral}
Table \ref{tab:rq1_gpt_neutral} shows GPT-4o authority bias results from Experiment 1, separated by neutral versus non-neutral response conditions. Authority bias emerges only when authoritative roles take clear positions, not when providing neutral responses. See Table \ref{tab:rq1_gpt_neutral} on page \pageref{tab:rq1_gpt_neutral} for detailed analysis.

\begin{table*}[hbp]
\centering
\footnotesize
\begin{tabular}{ll|c@{\;\;\;}c@{\;\;\;}c@{\;\;\;}c|c@{\;\;\;}c@{\;\;\;}c@{\;\;\;}c|c@{\;\;\;}c@{\;\;\;}c@{\;\;\;}c|c@{\;\;\;}c@{\;\;\;}c@{\;\;\;}c}
\toprule
& &\multicolumn{8}{c|}{\textbf{GPT-4o non-neutral option}} & \multicolumn{8}{c}{\textbf{GPT-4o neutral option}}\\
& &\multicolumn{4}{c|}{\textbf{FairEval}} & \multicolumn{4}{c|}{\textbf{Topical-Chat}} & \multicolumn{4}{c|}{\textbf{FairEval}} & \multicolumn{4}{c}{\textbf{Topical-Chat}} \\
&  &$t_1$& $t_4$ & $t_8$ & $t_{12}$ &$t_1$& $t_4$ & $t_8$ & $t_{12}$ &$t_1$& $t_4$ & $t_8$ & $t_{12}$ &$t_1$& $t_4$ & $t_8$ & $t_{12}$ \\

\midrule

Judge & $A_t$& 95.8 & 95.8 & 94.4 & 95.8 & 55.3 & 97.4 & 94.7 & 89.5 & 11.1 & 0.00 & 0.00 & 0.00 & 63.6 & 9.1 & 0.00 & 0.00\\

& $\kappa_t$& 0.89 & 0.89 & 0.86 & 0.89 & 0.38 & 0.95 & 0.90 & 0.80 & 0.00 & 0.00 & 0.00 & 0.00 & 0.00 & 0.00 & 0.00 & 0.00\\
\cmidrule{3-18}  
Foreman & $A_t$ & 80.0 & 93.3 & 98.3 & 98.3 & 48.7 & 92.3 & 87.2 & 84.6 & 35.0 & 0.00 & 0.00 & 0.00 & 42.9 & 4.8 & 28.6 & 9.5\\
& $\kappa_t$ & 0.60 & 0.84 & 0.96 & 0.96 & 0.31 & 0.84 & 0.75 & 0.70 & 0.00 & 0.00 & 0.00 & 0.00 & 0.00 & 0.00 & 0.00 & 0.00\\

\cmidrule{3-18}  
Management & $A_t$ & 87.7 & 100 & 98.2 & 100 & 39.4 & 100 & 100 & 97.0 & 21.7 & 0.00 & 0.00 & 0.00 & 63.0 & 7.4 & 11.1 & 7.4\\
&  $\kappa_t$ & 0.72 & 1.00 & 0.95 & 1.00 & 0.23 & 1.00 & 1.00 & 0.93 & 0.00 & 0.00 & 0.00 & 0.00 & 0.00 & 0.00 & 0.00 & 0.00\\

\midrule

Supervisor & $A_t$ & 83.6 & 100 & 98.4 & 96.7 & 45.5 & 90.9 & 88.6 & 90.9 & 10.5 & 0.00 & 0.00 & 0.00 & 62.5 & 6.3 & 0.00 & 12.5\\
&  $\kappa_t$ & 0.64 & 1.00 & 0.95 & 0.91 & 0.28 & 0.82 & 0.79 & 0.83 & 0.00 & 0.00 & 0.00 & 0.00 & 0.00 & 0.00 & 0.00 & 0.00\\
\cmidrule{3-18}  

Leader& $A_t$ & 78.9 & 96.5 & 96.5 & 96.5 & 45.0 & 92.5 & 95.0 & 87.5 & 21.7 & 0.00 & 0.00 & 4.3 & 50.0 & 10.0 & 10.0 & 15.0\\
& $\kappa_t$ & 0.60 & 0.92 & 0.92 & 0.92 & 0.27 & 0.85 & 0.90 & 0.77 & 0.00 & 0.00 & 0.00 & 0.00 & 0.00 & 0.00 & 0.00 & 0.00\\

\cmidrule{3-18}  
Mentor& $A_t$ & 78.9 & 100 & 100 & 96.5 & 33.3 & 78.6 & 76.2 & 85.7 & 21.7 & 0.00 & 4.3 & 0.00 & 50.0 & 5.6 & 11.1 & 5.6\\

& $\kappa_t$ & 0.61 & 1.00 & 1.00 & 0.92 & 0.18 & 0.61 & 0.57 & 0.71 & 0.00 & 0.00 & 0.00 & 0.00 & 0.00 & 0.00 & 0.00 & 0.00\\
\midrule

Specialist& $A_t$ & 81.0 & 98.3 & 96.6 & 94.8 & 41.7 & 97.2 & 94.4 & 91.7 & 18.2 & 0.00 & 4.5 & 9.1 & 54.2 & 0.00 & 0.00 & 4.2\\
&  $\kappa_t$ & 0.59 & 0.95 & 0.91 & 0.87 & 0.23 & 0.94 & 0.88 & 0.83 & 0.00 & 0.00 & 0.00 & 0.00 & 0.00 & 0.00 & 0.00 & 0.00\\
\cmidrule{3-18}  
Expert& $A_t$ & 85.2 & 100 & 98.4 & 100 & 48.7 & 94.9 & 97.4 & 89.7 & 21.1 & 0.00 & 0.00 & 0.00 & 52.4 & 4.8 & 4.8 & 14.3\\
& $\kappa_t$ & 0.71 & 1.00 & 0.96 & 1.00 & 0.32 & 0.90 & 0.95 & 0.80 & 0.00 & 0.00 & 0.00 & 0.00 & 0.00 & 0.00 & 0.00 & 0.00\\

\cmidrule{3-18}  
Attorney & $A_t$ & 82.1 & 100 & 100 & 96.4 & 33.3 & 97.0 & 93.9 & 100 & 8.3 & 0.00 & 0.00 & 4.2 & 44.4 & 7.4 & 7.4 & 0.00\\
& $\kappa_t$ & 0.63 & 1.00 & 1.00 & 0.91 & 0.19 & 0.94 & 0.88 & 1.00 & 0.00 & 0.00 & 0.00 & 0.00 & 0.00 & 0.00 & 0.00 & 0.00\\

\bottomrule

\end{tabular}
\caption{Comparison between two responses in GPT-4o, Experiment 1: Authority initially responds with \textit{Neutral} response versus \textit{Non-Neutral} response}
\label{tab:rq1_gpt_neutral}

\end{table*}

\section{Neutral response on Experiment 2}
\label{app:exp2_neutral}
Table \ref{tab:rq2_gpt_neutral} shows GPT-4o authority bias results from Experiment 2's content-controlled conversations, separated by neutral versus non-neutral response conditions. This experiment isolates role-based authority effects by controlling conversational content while varying only role labels. See Table \ref{tab:rq2_gpt_neutral} on page \pageref{tab:rq2_gpt_neutral} for detailed analysis.
\begin{table*}
\centering
\footnotesize
\begin{tabular}{ll|c@{\;\;\;}c@{\;\;\;}c@{\;\;\;}c|c@{\;\;\;}c@{\;\;\;}c@{\;\;\;}c|c@{\;\;\;}c@{\;\;\;}c@{\;\;\;}c|c@{\;\;\;}c@{\;\;\;}c@{\;\;\;}c}
\toprule
& &\multicolumn{8}{c|}{\textbf{GPT-4o non-neutral option}} & \multicolumn{8}{c}{\textbf{GPT-4o neutral option}}\\
& &\multicolumn{4}{c|}{\textbf{FairEval}} & \multicolumn{4}{c|}{\textbf{Topical-Chat}} & \multicolumn{4}{c|}{\textbf{FairEval}} & \multicolumn{4}{c}{\textbf{Topical-Chat}} \\
&  &$t_1$& $t_4$ & $t_8$ & $t_{12}$ &$t_1$& $t_4$ & $t_8$ & $t_{12}$ &$t_1$& $t_4$ & $t_8$ & $t_{12}$ &$t_1$& $t_4$ & $t_8$ & $t_{12}$ \\

\midrule

Judge & $A_t$& 81.6 & 100 & 100 & 98.0 & 62.2 & 97.3 & 100 & 100 & 96.8 & 32.2 & 0.00 & 0.00 & 65.2 & 4.3 & 4.3 & 13.0\\

& $\kappa_t$& 0.65 & 1.00 & 1.00 & 0.95 & 0.45 & 0.95 & 1.00 & 1.00 & 0.00 & 0.00 & 0.00 & 0.00 & 0.00 & 0.00 & 0.00 & 0.00\\
\cmidrule{3-18}  
Foreman & $A_t$ & 89.5 & 98.2 & 98.2 & 98.2 & 55.6 & 97.2 & 100 & 100 & 17.4 & 0.00 & 0.00 & 0.00 & 50.0 & 4.2 & 0.00 & 8.3\\
& $\kappa_t$ & 0.77 & 0.96 & 0.96 & 0.96 & 0.37 & 0.94 & 1.00 & 1.00 & 0.00 & 0.00 & 0.00 & 0.00 & 0.00 & 0.00 & 0.00 & 0.00\\

\cmidrule{3-18}  
Management & $A_t$ & 85.2 & 100 & 98.1 & 100 & 54.8 & 90.3 & 93.5 & 90.3 &11.5 & 3.8 & 0.00 & 0.00 & 55.2 & 3.4 & 6.9 & 3.4\\
&  $\kappa_t$ & 0.69 & 1.00 & 0.95 & 1.00 & 0.36 & 0.82 & 0.88 & 0.82 & 0.00 & 0.00 & 0.00 & 0.00 & 0.00 & 0.00 & 0.00 & 0.00\\

\midrule

Supervisor & $A_t$ & 87.9 & 94.8 & 96.6 & 98.3 & 45.0 & 85.0 & 90.0 & 90.0 & 4.5 & 0.00 & 0.00 & 0.00 & 50.0 & 5.0 & 10.0 & 5.0\\
&  $\kappa_t$ & 0.74 & 0.88 & 0.92 & 0.96 & 0.28 & 0.72 & 0.81 & 0.81 & 0.00 & 0.00 & 0.00 & 0.00 & 0.00 & 0.00 & 0.00 & 0.00\\
\cmidrule{3-18}  

Leader& $A_t$ & 89.7 & 100 & 100 & 96.6 & 40.0 & 100 & 97.5 & 95.0 & 18.2 & 9.1 & 4.5 & 0.00 & 65.0 & 0.00 & 5.0 & 10.0\\
& $\kappa_t$ & 0.78 & 1.00 & 1.00 & 0.93 & 0.25 & 1.00 & 0.95 & 0.90 & 0.00 & 0.00 & 0.00 & 0.00 & 0.00 & 0.00 & 0.00 & 0.00\\

\cmidrule{3-18}  
Mentor& $A_t$ & 91.7 & 100 & 100 & 100 & 47.7 & 97.7 & 90.9 & 93.2 & 20.0 & 5.0 & 0.00 & 0.00 & 68.8 & 0.00 & 0.00 & 0.00\\

& $\kappa_t$ & 0.81 & 1.00 & 1.00 & 1.00 & 0.32 & 0.96 & 0.83 & 0.87 & 0.00 & 0.00 & 0.00 & 0.00 & 0.00 & 0.00 & 0.00 & 0.00\\
\midrule

Specialist& $A_t$ & 86.5 & 98.1 & 100 & 100 & 47.4 & 92.1 & 100 & 92.1 & 14.3 & 3.6 & 0.00 & 3.6 & 68.2 & 13.6 & 9.1 & 13.6\\
&  $\kappa_t$ & 0.70 & 0.95 & 1.00 & 1.00 & 0.31 & 0.85 & 1.00 & 0.85 & 0.00 & 0.00 & 0.00 & 0.00 & 0.00 & 0.00 & 0.00 & 0.00\\
\cmidrule{3-18}  
Expert& $A_t$ & 83.9 & 100 & 100 & 100 & 54.5 & 87.9 & 93.9 & 93.9 & 20.8 & 0.00 & 0.00 & 0.00 & 59.3 & 7.4 & 7.4 & 3.7\\
& $\kappa_t$ & 0.70 & 1.00 & 1.00 & 1.00 & 0.38 & 0.77 & 0.88 & 0.88 & 0.00 & 0.00 & 0.00 & 0.00 & 0.00 & 0.00 & 0.00 & 0.00\\

\cmidrule{3-18}  
Attorney & $A_t$ & 83.9 & 98.4 & 98.4 & 98.4 & 44.4 & 97.8 & 95.6 & 95.6 & 5.6 & 5.6 & 5.6 & 0.00 & 53.3 & 0.00 & 13.3 & 13.3\\
& $\kappa_t$ & 0.68 & 0.96 & 0.96 & 0.96 & 0.29 & 0.95 & 0.91 & 0.91 & 0.00 & 0.00 & 0.00 & 0.00 & 0.00 & 0.00 & 0.00 & 0.00\\

\bottomrule

\end{tabular}
\caption{Comparison between two responses in GPT-4o, Experiment 2: Authority initially responds with \textit{Neutral} response versus \textit{Non-Neutral} response}
\label{tab:rq2_gpt_neutral}

\end{table*}

\end{document}